\documentclass[10pt,twocolumn,letterpaper]{article}

\usepackage[pagenumbers]{cvpr} 

\usepackage{graphicx}
\usepackage{amsmath}
\usepackage{amssymb}
\usepackage{booktabs}
\usepackage{xcolor}
\usepackage{comment}

\usepackage[pagebackref,breaklinks,colorlinks]{hyperref}

\usepackage[capitalize]{cleveref}
\crefname{section}{Sec.}{Secs.}
\Crefname{section}{Section}{Sections}
\Crefname{table}{Table}{Tables}
\crefname{table}{Tab.}{Tabs.}

\newcommand{\K}[0]{\mathbf{K}}
\newcommand{\R}[0]{\mathbf{R}}
\newcommand{\T}[0]{\mathbf{T}}
\newcommand{\x}[0]{\mathbf{x}}
\newcommand{\mbf}[1]{\mathbf{#1}}

\begin{document}

\title{DRaCoN -- Differentiable Rasterization Conditioned Neural Radiance Fields \\ for Articulated Avatars \vspace{-3mm}}


%
\author{Amit Raj$^{1,2}$\footnotemark \quad
Umar Iqbal$^1$ \quad
Koki Nagano$^1$ \quad
Sameh Khamis$^1$ \quad
Pavlo Molchanov$^1$ \quad \\
James Hays$^2$ \quad
Jan Kautz$^1$ \quad
\\ \quad 
\vspace{2mm}
$^1$NVIDIA Research\quad
$^2$Georgia Institute of Technology \quad \\
\url{https://dracon-avatars.github.io/} \\
}
%
%

\twocolumn[{%
\renewcommand\twocolumn[1][]{#1}%
\vspace{-2em}
\maketitle
\thispagestyle{empty}
\vspace{-1em}
\begin{center}
    \centering
    \includegraphics[width=0.90\linewidth,trim={0pt 15pt 0pt 0pt},clip]{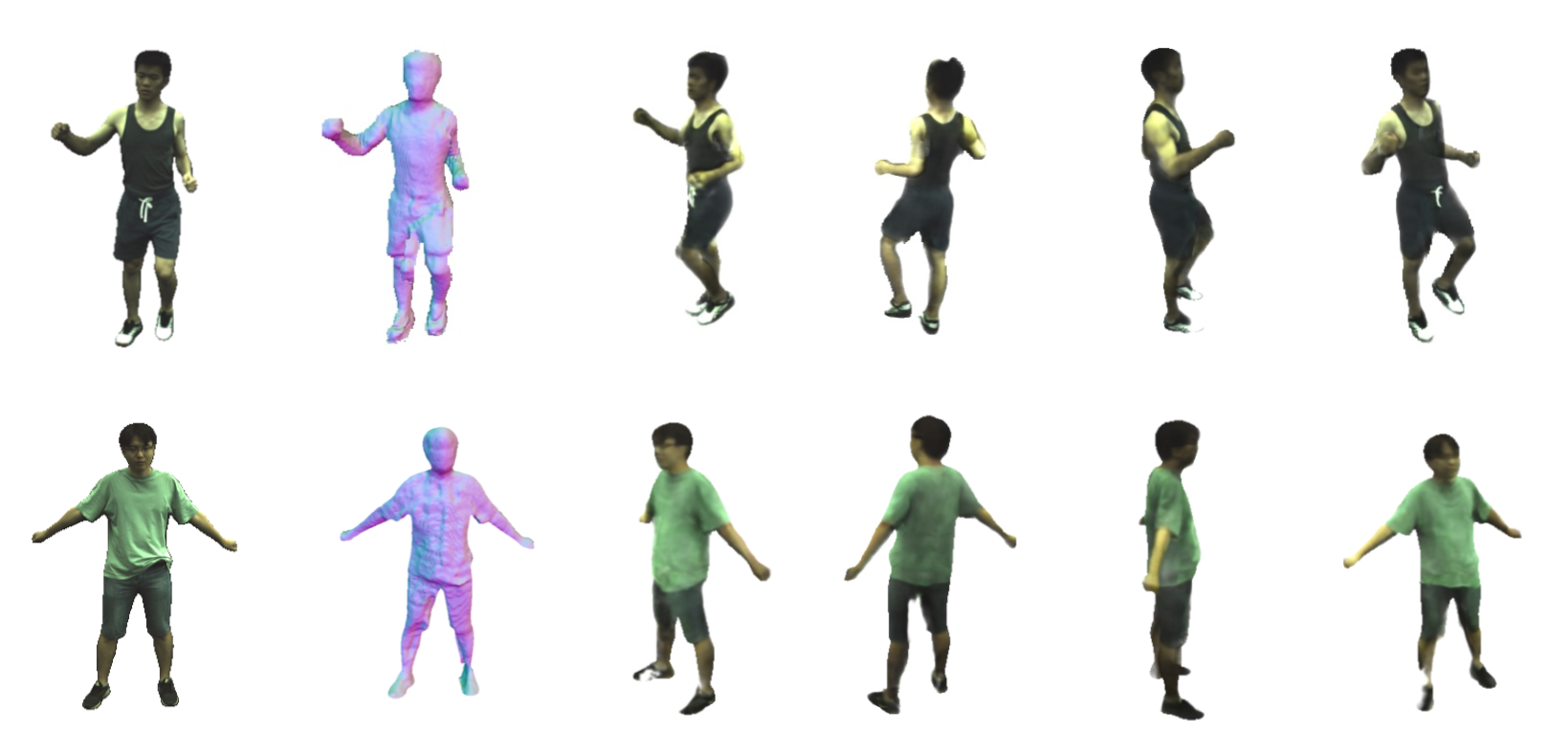}\vskip-2mm
    \captionof{figure}
    {
    We present, Differentiable Rasterization Conditioned Neural Radiance Fields (\textit{DRaCoN}), as an approach for learning a dynamic full body avatar from monocular or multi-view RGB videos. A training frame of an identity is shown in the first column. Our model provides articulation-based controls for the body model and allows novel view rendering in arbitrary poses. 
    }
    \label{fig:teaser}
\end{center}%
}]
\renewcommand{\thefootnote}{\fnsymbol{footnote}}
\footnotetext[1]{Work done during an internship at NVIDIA.}
\begin{abstract}
  Acquisition and creation of digital human avatars is an important problem with applications to virtual telepresence, gaming, and human modeling. Most contemporary approaches for avatar generation can be viewed either as 3D-based methods, which use multi-view data to learn a 3D representation with appearance (such as a mesh, implicit surface, or volume), or 2D-based methods which learn photo-realistic renderings of avatars but lack accurate 3D representations.
   In this work, we present, DRaCoN, a framework for learning full-body volumetric avatars which exploits the advantages of both the 2D and 3D neural rendering techniques. It consists of a Differentiable Rasterization module, DiffRas, that synthesizes a low-resolution version of the target image along with additional latent features guided by a parametric body model. The output of DiffRas is then used as conditioning to our conditional neural 3D representation module (c-NeRF) which generates the final high-res image along with body geometry using volumetric rendering. While DiffRas helps in obtaining photo-realistic image quality, c-NeRF, which employs signed distance fields (SDF) for 3D representations, helps to obtain fine 3D geometric details. Experiments on the challenging ZJU-MoCap and Human3.6M datasets indicate that DRaCoN outperforms state-of-the-art methods both in terms of error metrics and visual quality. 
\end{abstract}

\section{Introduction}


The rising interest in creating digital embodiments of real people for personalized gaming and immersive telepresence applications necessitates the creation of photorealistic full-body 3D digital avatars that are both scalable and low-cost.  
Creating controllable full-body avatars from monocular or multi-view videos of people in normal clothing is a challenging problem due to the wide range of plausible human motions, the lighting variations, and pose-dependent clothing deformations. 
Most recent approaches for relevant avatar generation can be categorized into: (i) 2D neural rendering methods, or (ii) 3D neural rendering methods\footnote{We follow the 2D and 3D neural rendering definitions by ~\cite{tewari2021advances}}. 

2D neural rendering methods use 2D or 2.5D guiding information to conditionally render images of avatars in novel views or poses~\cite{raj2020anr,prokudin2021smplpix,lwb2019,shysheya2019textured,wang2018vid2vid,nagano2018pagan}. These methods
rasterize some intermediate semantic labels or geometric representations such as UV-images, shapes, or poses and use 2D CNN-based generators (\ie, image translation network) to refine or render the final image from these intermediate representations. The main advantage of these methods is that they produce high-resolution photorealistic rendering results 
thanks to the high-resolution screen-space intermediate representations and high-capacity 2D CNN-based generators. 
However, these 2D methods show limited multi-view and temporal consistency in their rendering of novel views or poses despite requiring a large amount of training data, and do not learn the 3D representation of the object. 

3D neural rendering methods, on the other hand, aim to learn the 3D representation and appearance of the object or person which can be rendered in novel views and poses.  The state-of-the-art methods in this direction use implicit functions~\cite{saito2020pifuhd,raj2020pva} to model human avatars and use volumetric rendering for image generation~\cite{su2021anerf,peng2021neural,peng2021animatable,liu2021neural}. These methods not only provide 3D-consistent neural rendering but can also model pose-dependent deformations. 
While these implicit representations allow modeling the output as continuous functions, the complexity of the signals they can represent can be potentially limited when trying to model complex human body articulations and geometric variations, leading to less photorealistic outputs compared to the 2D-based neural rendering  methods~\cite{raj2020anr,prokudin2021smplpix,lwb2019,shysheya2019textured,wang2018vid2vid}.
Furthermore, these approaches which only rely on fully implicit architectures usually train a single network per identity and lack the strong generalization power seen in the 2D CNN-based generators. 

In this work, we propose Differentiable Rasterization Conditioned Neural Radiance Field (\textit{DRaCoN}), that leverages the advantages of both 2D and 3D neural rendering methods. Our hybrid approach not only provides view-consistent rendering but also achieves higher photorealism. 
DRaCoN learns a radiance field conditioned on pixel-aligned features that are obtained by differentiably rasterizing a high-dimensional latent neural texture for the target body pose and view using a parametric human body model, SMPL~\cite{loper2015smpl}. We enhance the rasterized neural texture with a 2D convolutional neural network which not only models the inter-part dependencies, but also allows us to render a low-resolution version of the target view before the volumetric rendering. We provide this low-resolution image as input to the above-mentioned conditional radiance field network with additional neural and geometric features, thereby, simplifying the task of learning 3D view synthesis. We also use the SMPL body model to warp the posed avatars into a canonical space. This allows us to generate avatars in arbitrary poses since the radiance fields are learned in the canonical space without having to allocate the capacity for modeling the body pose deformations. We demonstrate the effectiveness of our proposed approach on the challenging ZJU-MoCap and Human3.6M datasets. 

To summarize, our contributions are as follows:
\begin{enumerate}
    \item We present a novel hybrid framework for generating full-body, clothed, and dynamic avatars that combines screen-space and volumetric neural representations.
    \item Our framework learns identity-specific appearance space, allowing fine-grained dynamic texture synthesis during pose retargeting or change without having input images during inference. 
    \item We demonstrate the state-of-the-art performance on novel view and novel pose synthesis compared to recent volumetric approaches.
\end{enumerate}
\section{Related Work}

The existing works in learning articulated human avatars from RGB images or videos can be mainly classified into two categories -- 2D neural rendering-based methods and 3D neural rendering-based methods~\cite{tewari2021advances}.

\subsection{2D Neural Rendering Methods}
These methods aim to directly generate images of a person in novel views and poses using image-space rendering. Earlier methods treat the problem as an image-to-image translation task. Given an image of a target person with image representations of keypoints or dense meshes or semantic labels, these methods use an image translation model to directly render the target person in the style or pose of the source person possibly in different views~\cite{ma2017pose,neverova2018dense,raj2018swapnet,lassner2017generative,yang2020towards,zhu2019progressive,esser2018variational,pumarola2018unsupervised,yoon2021poseguided,wang2018vid2vid,chan2019everybody}. 
The main limitation of these methods is that they lack multi-view and temporal consistency under significant viewpoint and pose changes since these 2D-based method generally do not learn any notion of 3D space. 
Instead of fully relying on the image-space translation network, Liquid Warping GAN~\cite{lwb2019,liu2021liquid} uses UV-correspondences between the source and target meshes to explicitly warp the source image to the target pose and then uses an additional network to refine the warped image. Given a target pose in form of 2D keypoints, Huang~\etal~\cite{huang2021few} predict the UV-coordinates of a dense mesh in image space and then generate the target image by sampling from a learned UV texture map. The main limitation of these methods is that they require some exemplar images to be available during inference, limiting their semantic editing capability, such as fine-grained control of the target appearance. 

Textured Neural Avatar~\cite{shysheya2019textured} removes the need of exemplar images but relies on multi-view data for training. Articulated Neural Renderer (ANR)~\cite{raj2020anr} builds on a similar idea but instead uses reconstructed 3D meshes to rasterize a learned neural texture on image space from which the final rendered image is obtained using a convolutional network. While these approaches show impressive results and have the capability to generate photorealistic images, their output is limited to 2D images only and 
does not provide 3D representations of the person.  Inspired by these methods, we propose a method that not only renders photorealistic images but also learns 3D representation of the person, allowing view-consistent rendering under large view or pose changes.  

\subsection{3D Neural Rendering Methods.}
These methods aim to create geometric representations of avatars as well as their 3D appearance that can be rendered in novel viewpoints and poses.\\
\textbf{Mesh-Based Approaches.} 
Given a monocular or multi-view video, earlier methods reconstruct the detailed geometry and textures using parametric mesh fitting~\cite{Alldieck_2018_CVPR,Alldieck_2019_CVPR,zhi2020texmesh,deepcap}. Alldieck~\etal~\cite{Alldieck_2018_CVPR} fit the SMPL body model to all frames and  optimize for per-vertex offsets using person silhouette information to capture the clothing and hair details. A detailed texture map is then generated by back-projecting the image colors to all visible vertices. The novel views and poses of the person can  be easily generated by articulating the SMPL mesh and rendering it to the image with the obtained texture map. The methods~\cite{Alldieck_2019_CVPR,zhi2020texmesh} adopt a similar strategy but replace the optimization-based framework used in~\cite{Alldieck_2018_CVPR} with learning-based components for faster processing. In all cases, they produce
a parametric mesh with a texture map which can be articulated and rendered in a new view. However, the main limitation of these methods is that they rely on a fixed mesh topology which cannot accurately capture complex clothing and hair geometries. Also, the articulated meshes do not account for pose-conditioned geometric deformations and changes in texture. Hence, the rendered results lack realism and high-frequency details. In contrast, our volumetric representation directly learns these details that are not captured by the parametric models. 
\\
\textbf{Implicit Approaches.} More recently, PIFu and PIFuHD \cite{Saito2019,Saito2020} proposed to use implicit functions and pixel-aligned features for human reconstruction from single or multi-view RGB images. Since the implicit functions represent the body's surface in a continuous 3D space, these methods can reconstruct highly-detailed human body shape. However, the reconstructed geometries are not rigged and cannot be articulated in novel poses. To this end, ARCH~\cite{huang2020arch} and S3~\cite{yang2021s3} also reconstruct meshes that can be articulated with novel poses. However, similar to other mesh-based models, rendering the reconstructed bodies in novel poses does not account for pose-dependent deformations. Scanimate \cite{saito2021scanimate} learns to fit a deformable implicit function of surface data to RGBD scans, allowing for additional articulation but requires 3D supervision to train.

The most recent methods take inspiration from NeRF~\cite{mildenhall2020nerf} or its variants \cite{lin2021barf,yu2021plenoctrees,barron2021mip,oechsle2021unisurf,wang2021neus} and represent human avatars using pose-conditioned implicit 3D representations~\cite{tiwari2021neural,peng2021animatable,peng2021neural,kwon2021neural,liu2021neural}. We refer the readers to the comprehensive report on advances in neural rendering \cite{tewari2021advances} for additional insight about these approaches. These methods not only allow photorealistic novel view synthesis but also provide detailed human body geometry with pose dependent variations. NeuralBody~\cite{peng2021neural} exploits the SMPL body model and learn latent codes corresponding to each vertex. These latent codes are then diffused using a sparse convolution module and sampled according to the target body pose and used to condition the NeRF model. A-NeRF~\cite{su2021anerf} and AnimatableNeRF~\cite{peng2021animatable} use body pose information to canonicalize the sampled rays and learn neural radiance fields in the canonical space, which helps the learned avatar to generalize across different poses.
Our hybrid framework leverage advantages of multiple representations: the parametric model provides articulation-based controls and promotes generalization through learnable neural textures enhanced by a 2D CNN-based generator, while our NeRF enables conditional 3D view synthesis and high-fidelity 3D geometry extraction. 
\begin{figure*}[t]
    \centering
    \includegraphics[width=1.0\linewidth,trim={0cm 1.5cm 0cm 0cm},clip]{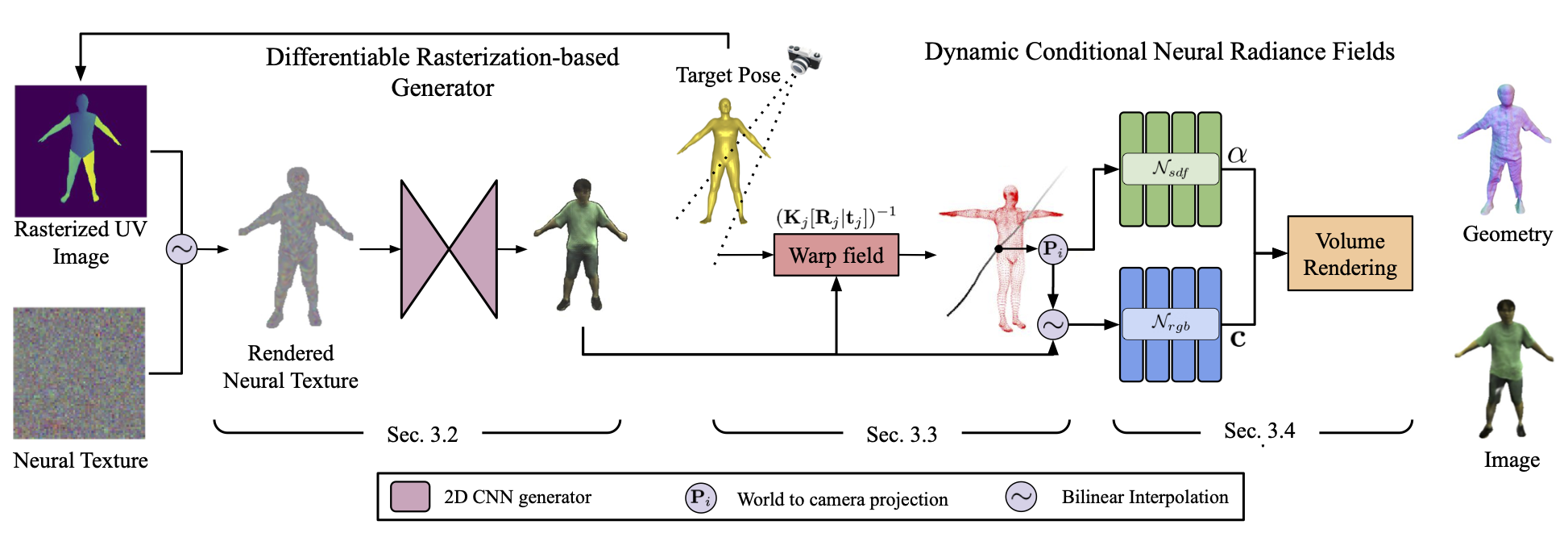}
    \caption{Overview of the proposed approach. Given a target camera and a parametric body model of an arbitrary pose, DRaCoN \emph{i)} employs a differentiable rasterizer (DiffRas) with identity-specific neural textures to rasterize pixel-aligned neural features for the target view; \emph{ii)} warps the 3D rendering points from the posed space to canonical volume using a learned warping field and samples neural features;   \emph{iii)} uses the rendered pixel-aligned features as input to the conditional neural 3D representation module (c-NeRF) which produces the final image using volumetric rendering. We can easily extract fine surface geometry as the zero-level set of the predicted signed distance values.}
    \label{fig:framework}
    \vspace{-0.5cm}
\end{figure*}

\vspace{-0.3cm}
\section{Method}
Our goal is to learn a controllable full-body avatar that can synthesize geometrically consistent images of the person in novel views using 3D neural rendering, while also leveraging the advantages of 2D neural rendering for photorealism. For training, we assume a set of calibrated multi-view or monocular videos is available. We also assume that the human bodies in the training video frames have been tracked using the SMPL~\cite{loper2015smpl} body model. An overview of our approach can be seen in Figure~\ref{fig:framework}. Our framework consists of three modules: \emph{differentiable rasterization (DiffRas)} (Section~\ref{sec: DiffRas}), \emph{WarpField} (Section~\ref{sec: WarpField}) module and \emph{conditional-NeRF} module (Section~\ref{sec: GeoNet}). Given a target pose represented as the parameters of SMPL body model, DiffRas rasterizes high-dimensional identity-specific neural features in the target view. These pixel-aligned features are then used as a condition during 3D neural rendering. To simplify the learning of otherwise highly dynamic human bodies, the WarpField module maps the sampled 3D points in 3D world space to a canonical space. Finally, the canonicalized 3D points and pixel-aligned features are fed as input to the conditional neural 3D representation module (c-NeRF) which generates the final high-res image and detailed 3D surface geometry using volumetric rendering. 
In the following, we explain each of these modules in greater details. 

\vspace{-0.3cm}
\subsection{Preliminaries}

Given a set of posed images $ \{ \mathbf{I}_i  \in \mathbb{R}^{{H}\times{W}\times3} \}_{i=1}^N$ with their associated intrinsic  
$\{\K_i\}_{i=1}^N$ and extrinsic camera parameters (rotation $\{\R_i\}_{i=1}^N$ and translation $\{\mbf{T}_i\}_{i=1}^N$), we define the world to camera projection  as 
$\mathbf{P}_i{=}\K_i [ \R_i | \T_i ]$. We also assume that the pose and shape parameters $(\theta_i \in \mathbb{R}^{24\times3}, \beta_i \in \mathbb{R}^{10})$ of the tracked SMPL meshes for each frame $i$ are available from which the 3D mesh vertices $\mathbf{V}$ can be obtained using a linear function $\mbf{V}_i = \mathcal{M}(\theta_i, \beta_i)$ as defined in~\cite{loper2015smpl}.

NeRF~\cite{mildenhall2020nerf} learns a function  $\mathcal{N} (\x,\mbf{d}) : \mathbb{R}^{N_{\x} \times N_{\mbf{d}}} \rightarrow \mathbb{R}^4$ which takes the positional encodings of a world space position $\x$ and viewing direction $\mbf{d}$ as input and returns the radiance $\mbf{c} \in \mathbb{R}^3$ and associated differential opacity $\alpha \in \mathbb{R}$ at that location:
\[
\mbf{c},\mbf{\alpha} = \mathcal{N}(\gamma_L(\x),\gamma_M(\mbf{d})), 
\]
where $\gamma_L(\x)$ and $\gamma_M{(\mbf{d})}$ are the positional encodings of $\x$ and $\mbf{d}$ with $L$ and $M$ octaves, respectively,  as described in \cite{Mildenhall2019}. For ease of readability we drop the dependence on $\mbf{d}$ in all future discussion.

The NeRF network is trained by tracing rays into the scene from a camera origin through each pixel of the image. For image $i$ the $j$th pixel location is given by $u^j_i\in\mathbb{R}^2$ and the camera origin $\mbf{o} \in \mathbb{R}^3$ and ray direction $\mbf{d} \in \mathbb{R}^3$ are calculated as follows:
\begin{equation}
\mbf{o}_i = -\R_i^{-1}\T_i,
\end{equation}
\vspace{-0.3cm}
\begin{equation}
\mbf{d}^j_i = \frac{\mbf{P}^{-1} u^j_i - \mbf{o}_i}{|| \mbf{P}^{-1} u^j_i - \mbf{o}_i||}.
\end{equation}
We then sample points uniformly along the ray given by
\[
\mbf{x}^j_i (t) = \mbf{o}_i + t \mbf{d}^j_i.
\]
We drop the parameterization variable $t$ in all subsequent equations for notational simplicity.

\vspace{-0.2cm}
\subsection{Differentiable Rasterization (DiffRas)}
\label{sec: DiffRas}
Our goal is to generate pixel-aligned features $\mbf{F}~{\in}~\mathbb{R}^{H{\times}W{\times}D}$ capturing the geometric and appearance properties of the person in the target viewpoint and body pose. In order to make sure that these features are consistent across different body poses and views, we define them in the canonical UV-space of the SMPL body model and rasterize them with the 3D mesh. Specifically, we learn a set of D-dimensional (D=128) neural features $\mbf{Z}_i \in \mathbb{R}^{M \times M \times D}$ where the spatial coordinates correspond to the UV-coordinates of SMPL. Given the SMPL vertices $\mbf{V_i}$, DiffRas differentiably rasterizes these neural features to the image space: $\texttt{DiffRas}(\mbf{Z}, \mbf{V}_i) \rightarrow \mbf{F_i}$. 
The learnable neural features $\mbf{Z}$ are optimized end-to-end using the image reconstruction losses over all training frames.   

Unlike previous works which rely on pixel-aligned features from exemplar images,  \cite{raj2020pva,yu2021artificial}, the addition of the differentiable rasterizer eschews the need for input images during inference as the pixel-aligned features can now be rasterized on the fly for any arbitrary viewpoint for any identity. In practice, we enforce the first three channels of $\mbf{F_i}$ to be a low resolution version the target image. This allows to exploit the latest advances in 2D neural rendering, for example, by using photometric and perceptual losses on the generated images to improve photo-realism. Also, since the features are already rasterized in the target pose, it simplifies the problem for subsequent components of our approach. In contrast to existing methods~\cite{peng2021animatable,peng2021neural,su2021anerf} which encode all appearance and geometric details in the NeRF module using a highly non-linear mapping, our approach learns these details in a well-defined UV-space thereby simplifies neural avatar learning. 

\vspace{-0.1cm}
\subsection{Warping Fields}
\label{sec: WarpField}

 
 The NeRF model as originally proposed in~\cite{mildenhall2020nerf} is trained for static scenes only. To simplify learning of the complex articulated geometries of human bodies, we map sampled points $\mbf{x}$ in posed space to canonical space before passing them to c-NeRF using a warping field $\mathcal{W}$.   
 Particularly, each point $\mbf{x}_i^j$ in posed space is mapped to point $\hat{\mbf{x}} = \mathcal{W}(\mbf{x})$ in the canonical space. 
 
 However, as shown in \cite{lombardi2019neural}, learning the warping field and $\mathcal{N}(\bold{x})$ simultaneously is an under-constrained problem. To this end, we exploit the tracked SMPL meshes to regularize the learning of the warping field. 
 %
 Specifically, let $G_k(\theta)$ represent the local transformation of joint $k$ with respect to its parent in the SMPL body model, we define the transform to warp a canonical vertex $\hat{\mbf{v}_l} \in \mathbb{R}^3$ to posed space as  follows:

\begin{equation}
\mathcal{\phi}(\mbf{v}_l)= \sum_{m=1}^J w_m(\mbf{v}_l) (\prod_{k \in \mathcal{K}_m} G_k(\theta)),
\end{equation}
where $\mathcal{K}_m$ is the kinematic chain for joint $m$ and $w_m(\mbf{v}_l)$ are the skinning weight associated with vertex $\mbf{v}_l$. 
The location of the posed vertex is obtained as: $ \mbf{v}_l = \phi(\hat{\mbf{v}}_l) \hat{\mbf{v}}_l $.
 
 We use this skinning field to initialize the warping field $\mathcal{W}$. Specifically, for any point $\x_i^j$ in posed space, we find the vertex $\mathbf{v} \in \mbf{V}_i$ that is closest to this point, and use the associated skinning weights to canonicalize $\x_i^j$:
 \begin{equation}
 p(\x) = \min_{\mbf{v} \in \mbf{V}} || \x - \mbf{v} ||, 
 \end{equation}
 \begin{equation}
\hat{\x} = \mathcal{W}(\x) = \phi^{-1}(p(\x))\x. 
 \end{equation}

 \noindent\textbf{Learning Residuals for Warp Field.}
 The skinning weights associated with the naked body model do not account for the geometry outside the parametric mesh or pose-based geometric deformation of clothing. To account for these variations, we learn a residual field $\Delta \mathcal{W}(\x,\mbf{f}_i^{\x}) : \mathcal{R}^{3+d} \rightarrow \mathcal{R}^3$ and obtain the final canonicalized point $\hat{\x}^c$ as:
 \[
 \hat{\x}^c = \hat{\x} + \Delta  \mathcal{W}(\x, \mbf{f}_i^{\mbf{x}}).
 \]
Here $\mbf{f}_i^{\x} \in \mbf{F}_i$ are the pixel-aligned feature associated with point $\x$ obtained from the differentiable rasterizer (Section~\ref{sec: DiffRas}). 

\subsection{Conditional 3D Representation}
\label{sec: GeoNet}
The DiffRas module produces a low resolution image with no associated geometric information. To this end, we learn a conditional neural 3D representation (c-NeRF) model that takes as input $\hat{\x}^c$ and pose-specific features $\mbf{f}_i^{\mbf{x}}$, and produces a high-fidelity 3D surface geometry as well as generates a higher resolution version of the image. We follow ~\cite{wang2021neus,yariv2021volume} and represent the avatar using two functions $\mathcal{N}_{sdf}$ and $\mathcal{N}_{color}$ responsible for modeling surface geometry and texture, respectively. The neural signed distance function (SDF) $\mathcal{N}_{sdf} : \mathbb{R}^{L+D} \rightarrow \mathbb{R}$ takes as input the positional encoding of the canonicalized point $\hat{\mbf{x}}^c$ and pose-specific features $\mbf{f}_i^{\mbf{x}}$ and produces the signed distance of the point \wrt body surface geometry. Following \cite{yariv2021volume}, we generate the differential opacity at the point from the SDF using the following function:
\[
\alpha = \Psi(\mathcal{N}_{sdf}(\hat{\x}^c,\mbf{f}_i^{\mbf{x}}), \beta). 
\]
where $\Psi$ is the cumulative distribution function (CDF) of a lapalacian distribution with scale $\beta$ which is a learnable parameter.
The function $\mathcal{N}_{rgb}$ also takes $\hat{\x}^c$ and pose-specific features $\mbf{f}_i^{\mbf{x}}$ and produces the color, $\mbf{c}=\mathcal{N}_{rgb}(\hat{\x}^c,\mbf{f}_i^{\mbf{x}})$, associated with the point. To better inform the input, we also concatenate the UV-coordinates of the sampled features as an additional information. The final RGB image is generated using volume rendering and the 3D geometry is obtained as the the zero-level set of the SDF. 

Note that using SDF to represent body surface has the advantage that it is interpretable and the geometry can be extracted easily without the need of a careful selection of threshold for density values as used in existing NeRF-based neural avatar methods~\cite{peng2021neural,peng2021animatable,su2021anerf}.

\subsection{Training and Losses}

 \noindent \textbf{Texture Branch:}
 The c-NeRF network is trained with a photometric reconstuction loss for the color $C(u_i^j)$ accumulated along each ray w.r.t ground truth color $\hat{C}(u_i^j)$:
 \[
 \mathcal{L}_\text{recon} = ||C(u^j_i) - \hat{C}(u_i^j)||_2. 
 \]
 We also find that the warping module can ``cheat" by generating large $\Delta \x$ values that place the sampled points at arbitrary locations to help optimize the learning. This causes generalization to suffer as the mapping for novel poses become uncertain. We constrain the norm of the deformation to limit the maximal displacement that the warping module can produce:
 \[
 \mathcal{L}_\text{delta} = ||\Delta \mathcal{W}(\x,\mbf{f_i}^{\x})||_2
 \]
 The c-NeRF network is then optimized as follows:
 \begin{equation}
     \mathcal{L}_\text{nerf} =  \lambda_\text{recon} \mathcal{L}_\text{recon} +  \lambda_\text{delta} \mathcal{L}_\text{delta}
 \end{equation}
 \textbf{DiffRas Branch:}
While there is no exact supervision for the output of DiffRas module $\mathbf{F}$, we constrain the first three channels of $\mathbf{F}$ as in ANR~\cite{raj2020anr} as follows:
\[
\mathcal{L}_\text{pixel} = ||\mbf{F}_i[:3] - \mbf{I}_{i}||.
\]
\[
\mathcal{L}_\text{vgg} = \sum_{l} w_l ||\phi_l(\mbf{F}_i)[:3] - \phi_l(\mbf{I}_{i})||.
\]
Where, $\phi_l$ return features at different layers of a pretrained VGG-network \cite{simonyan2014very}. Additionally, we constrain the fourth channel to learn the mask of the generated identity. 
\[
\mathcal{L}_\text{mask} = \textrm{BCE}(\mbf{F}_i[4] , \mbf{M}_i).
\]
The total loss for the DiffRas branch is then computed as:
\begin{equation}
    \mathcal{L}_\text{DiffRas} = \lambda_\text{pixel} \mathcal{L}_\text{pixel} + \lambda_\text{vgg} \mathcal{L}_\text{vgg} + \lambda_\text{mask} \mathcal{L}_\text{mask}.
\end{equation}

\noindent\textbf{Geometry Branch:}
The geometry network, $\mathcal{N}_{sdf}$, is encouraged to learn SDF by using the eikonal regularization\cite{icml2020_2086}:
\begin{equation}
    \mathcal{L}_\text{eik} = ||\Delta \mathcal{N}_{sdf}(\x,\mbf{f^x}) - 1||. 
\end{equation}
Additionally, to encourage smoothness of the learnt geometry we apply a Total Variation (TV) loss over the estimated density as follows:
\begin{equation}
    \mathcal{L}_\text{TV} = ||\Psi(\x) - \Psi(\x+\delta \x)||.
\end{equation}
Furthermore, following Sitzmann et al.\cite{sitzmann2020implicit}, we apply exponential regularization for encouraging non-zero SDF at almost all locations except the surface:
\begin{equation}
    \mathcal{L}_\text{exp} = \mathbb{E}_\x[\exp{(-\alpha * \mathcal{N}_{sdf}(\x))}]
\end{equation}
where $\alpha >> 1$. Finally, we also use an SMPL guided SDF regularization, $\mathcal{L}_{face}$, for the face region that encourages SDF of the samples around face regions to be close to SDF w.r.t to SMPL mesh. The regularization for the geometry is then given by:
\begin{equation}
    \mathcal{L}_\text{geom} = \lambda_\text{eik} \mathcal{L}_\text{eik} + \lambda_\text{TV}\mathcal{L}_{TV} + \lambda_\text{exp} \mathcal{L}_\text{exp} +  \lambda_\text{face} \mathcal{L}_\text{face}
\end{equation}

The training objective for our framework is thus given by
\begin{equation}
    \mathcal{L}_\text{DRaCoN} = \underbrace{w_\text{nerf} \mathcal{L}_\text{nerf}}_\textrm{NeRF objective} + \underbrace{w_\text{DiffRas} \mathcal{L}_\text{DiffRas}}_\textrm{DiffRas objective} + \underbrace{w_\text{geom} \mathcal{L}_\text{geom}}_\textrm{Geometry Regularization}
\end{equation}

 \subsection{Progressive Growing}
 In practice, we first train both the DiffRas and c-NeRF at a low resolution and then train c-NeRF at successively higher resolutions while keeping the resolution of DiffRas constant. This progressive training accelerates the convergence since the network mostly needs to focus on the coarse structures early in the training. For training DiffRas, we also calculate a per-frame affine transform that crops the image in the scene to a tight 256 $\times$ 256 image. This ensures that the DiffRas always sees similarly sized images regardless of where the actor is in the world space.

\section{Experiments}

\begin{figure*}[hbt!]
  \centering
    \setlength{\tabcolsep}{1pt}
\scalebox{0.95}{
  \begin{tabular}{cccccccccc}
  \multicolumn{10}{c}{\includegraphics[trim={0cm 0cm 0cm 0cm},clip, width=\textwidth]{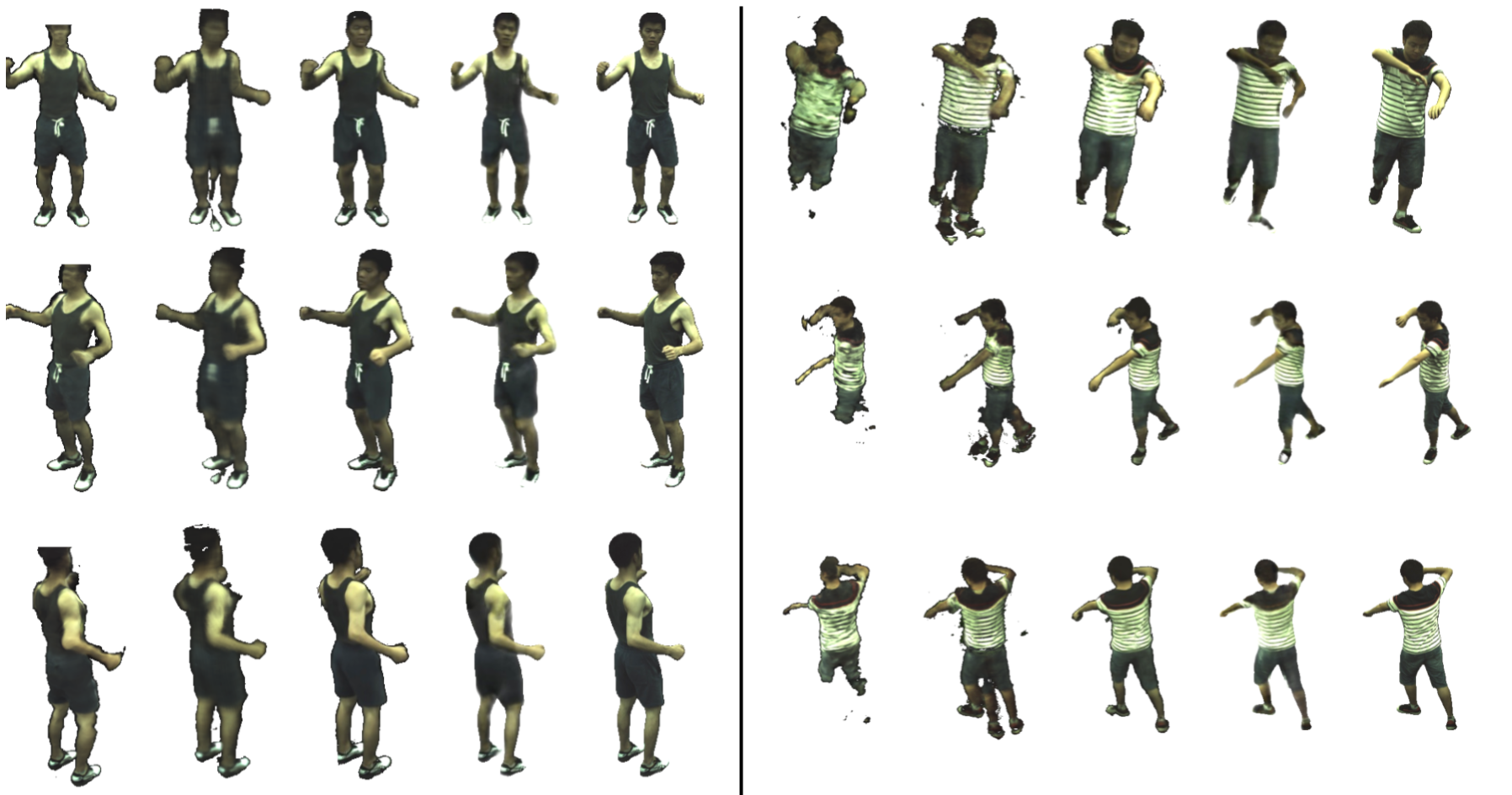}} \\
  ~~~A-NeRF~~~  & Animatable & Neural & ~~~~Ours~~~~~  & ~~Ground~~ & ~~~~~~~~~~A-NeRF~~~  & Animatable~ & Neural & ~~~~~Ours~~~~  & ~~~~Ground~~~~  \\ 
    \cite{su2021anerf}                         & ~NeRF~\cite{peng2021animatable}       & ~~Body\cite{peng2021neural}   &           &  Truth             &   ~~~~~~~\cite{su2021anerf}      & NeRF~\cite{peng2021animatable}       & ~Body~\cite{peng2021neural}   &      & Truth
  \end{tabular}
}
\caption{Comparison of novel view and pose synthesis. In contrast to the state-of-the-art, our proposed approach yields better textures quality owing to conditional features from DiffRas. Notably, our approach also produces better person shape. }
\label{fig:zjuc}
\end{figure*}


\begin{figure}
  \centering
    \setlength{\tabcolsep}{10pt}
\scalebox{0.48}{
  \begin{tabular}{c c c c c c c c}
  \multicolumn{8}{c}{\includegraphics[trim={0cm 0cm 0cm 0cm},clip, width=\textwidth]{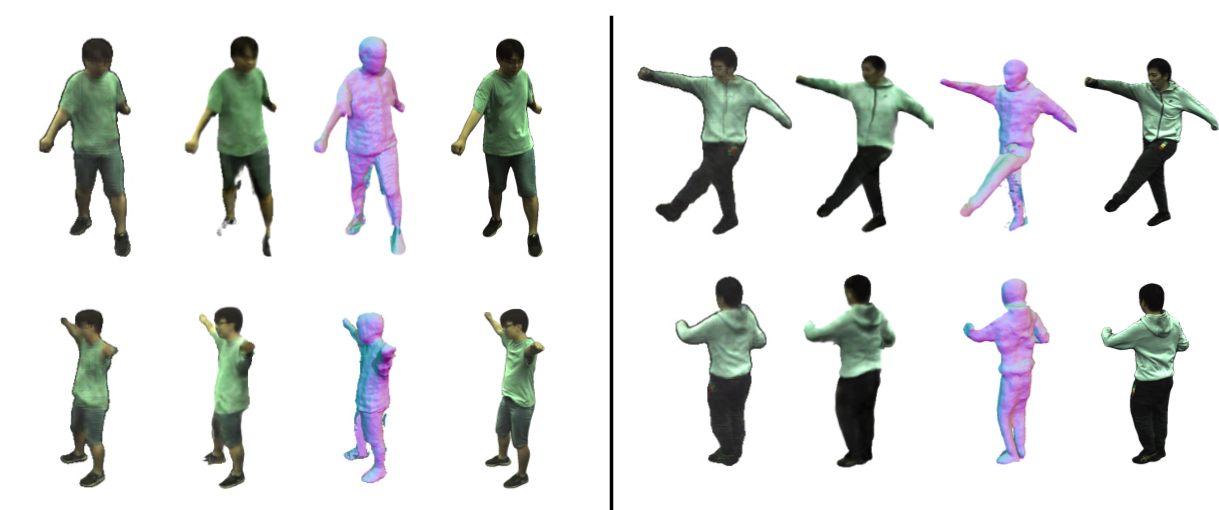}} \\
  ~~ ~ ~ ~DiffRas & c-NeRF ~ & ~~ Geometry & ~~ GT ~~~~~~ & ~~~~~ DiffRas ~~ & c-NeRF ~~ & Geometry & GT
  \end{tabular}
}
\caption{Novel view (above) and novel pose (below) results of each stage of our model. We see that the c-NeRF module adds necessary details using intermediate output from the DiffRas stage. The geometry is obtained by sampling $\mathcal{N}_{sdf}$ using a uniform grid and running marching cubes over it.}
\label{fig:anerf}
\end{figure}

\begin{figure}
  \centering
    \setlength{\tabcolsep}{1pt}
    \renewcommand{\arraystretch}{0.1}
\scalebox{0.49}{
  \begin{tabular}{ccccc}
  \multicolumn{5}{c}{\includegraphics[trim={0cm 0cm 0cm 0cm},clip, width=\textwidth]{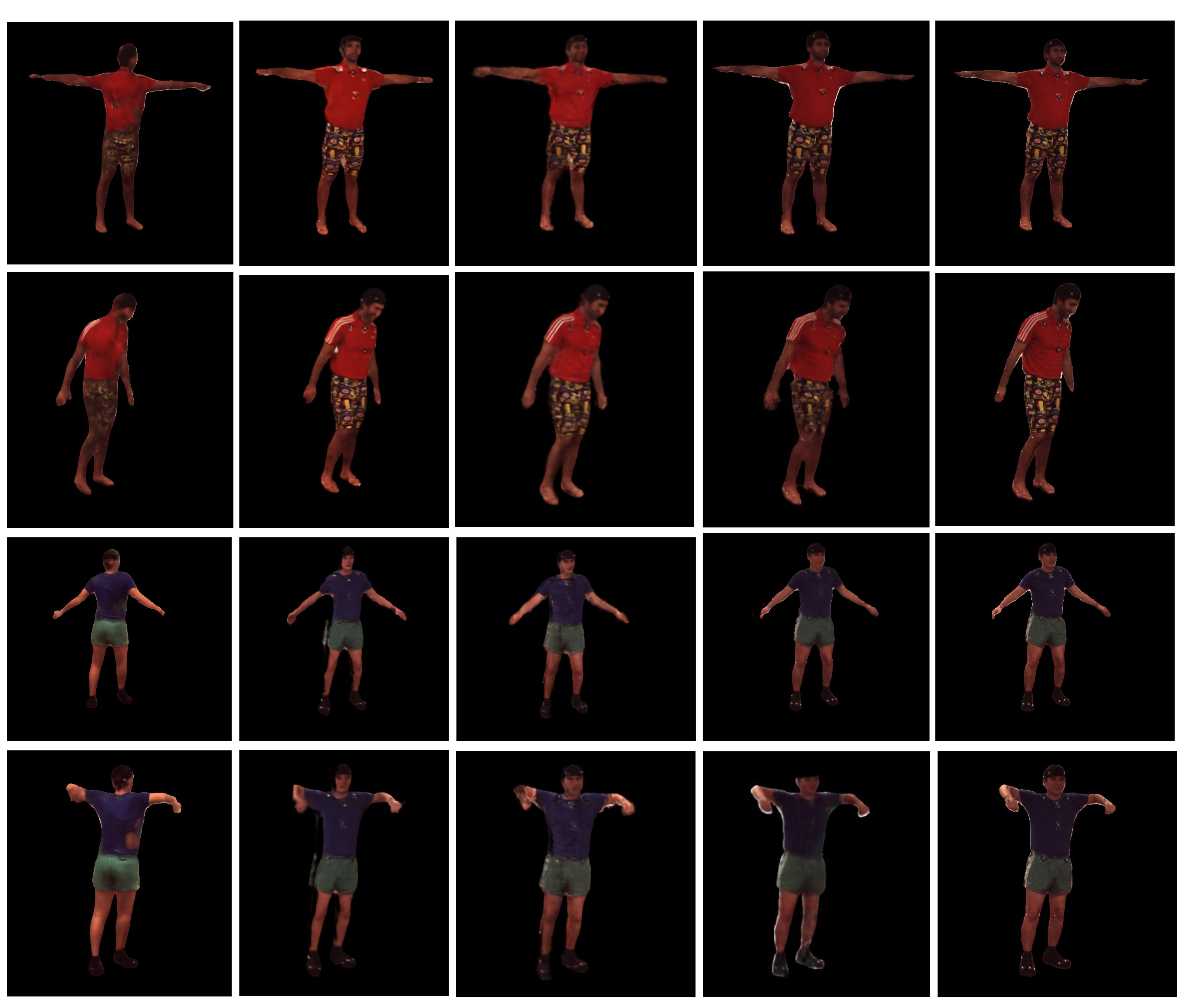}} \\
  ~~~~~~~~~~~NHR\cite{wu2020multi}~~~~~~~ & ~~~~~~~~~NeuralBody\cite{peng2021neural}~~~~~~~ & Animatable-NeRF~\cite{peng2021animatable}~~~~~~~ & ~~~~~~~~~~~Ours~~~~~~~ & ~~~~~~~~GT~~~~~~
  \end{tabular}
}
\caption{Comparison of novel view (odd rows) and novel pose (even rows) synthesis w.r.t NHR, NeuralBody, and Animatable-Nerf. Our model better preserves the shape of the actor under novel poses.}
\label{fig:comparison}
\end{figure}

\begin{table*}[h]
\small
    \centering
    \begin{tabular}{l|cccc | cccc | cc}
    &\multicolumn{4}{c}{ZJU-315} & \multicolumn{4}{c}{ZJU-377} & \multicolumn{2}{c}{User Study} \\
    \specialrule{0.12em}{0.05em}{0.05em}
     & SSIM $\uparrow$ & PSNR $\uparrow$ & \multicolumn{2}{c}{LPIPS $\downarrow$} & SSIM $\uparrow$ & PSNR $\uparrow$ & \multicolumn{2}{c}{LPIPS $\downarrow$} & Num Votes & Mean preference \\
     &&& Alexnet & VGG &&& Alexnet & VGG \\
     \specialrule{0.12em}{0.05em}{0.05em}
    A-NeRF~\cite{su2021anerf}  &  $0.930$ & $17.711$ & $0.094$ & $0.073$ & $0.927$ & $16.528$ & $0.101$ & $0.077$ & $53$ & $7.5 \pm 1.5 \%$  \\
    Anim-NeRF~\cite{peng2021animatable}  &  $0.917$ & $16.774$ & $0.101$ & $0.083$ & $0.940$ & $17.971$ & $0.075$ & $0.066$ & $15$ & $2.5 \pm 0.64 \%$ \\
    NeuralBody~\cite{peng2021neural} &   $\mathbf{0.947}$ & $18.608$ & $0.092$ & $0.090$ & $\mbf{0.950}$ & $20.072$ & $0.071$ & $0.060$ & $168$ & $24 \pm 3.25 \%$   \\
    \hline
    Ours &  $0.922$ & $\mbf{21.315}$ & $\mbf{0.053}$ & $\mbf{0.046}$ & $0.946$ & $\mbf{21.084}$ & $\mbf{0.048}$ & $\mbf{0.044}$ & $464$ & $\mbf{66} \pm 4 \%$ 
    \end{tabular}
    \caption{Novel pose synthesis results on ZJU-MoCap dataset. Our approach outperforms recent baselines on most supervised metrics.}
    \vspace{-5mm}
    \label{tab:comparison}
\end{table*}

\begin{figure}
  \centering
    \setlength{\tabcolsep}{1pt}
    \renewcommand{\arraystretch}{0.1}
\scalebox{0.49}{
  \begin{tabular}{cccccc}
  \multicolumn{6}{c}{\includegraphics[trim={0cm 0cm 0cm 0cm},clip, width=\textwidth]{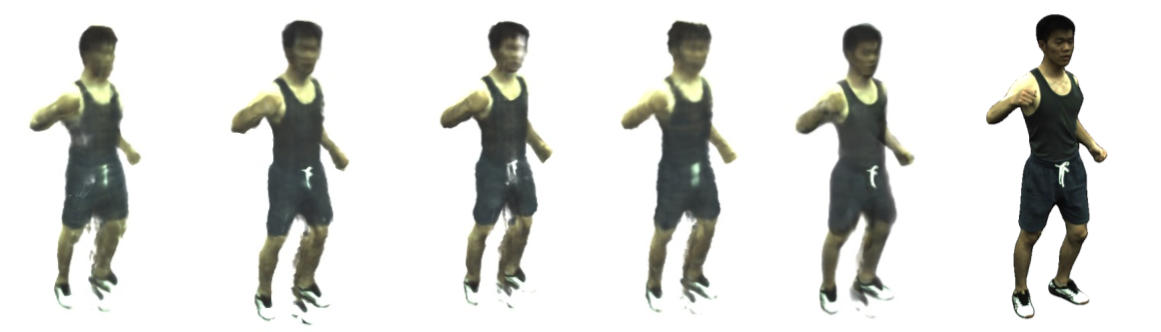}} \\
  Ours$(-\mathcal{L}_{\mathrm{\mathrm{DiffRas}}}) ~~~~~$ & ~~Ours$(-\mathcal{L}_{\mathrm{\mathrm{face}}})$~~~~~~~~ & ~~Ours$(-\mathcal{L}_{\mathrm{eik}})$~~~~~~~~~ & ~~Ours$(-\mathcal{L}_{\mathrm{\mathrm{delta}}})$~~~~~~~~ & ~~~~~~~Ours & ~~Ground Truth
  \end{tabular}
}
\caption{We demonstrate the importance of each module in our framework by comparing against models trained under the following settings: (a) without $\mathcal{L}_{\mathrm{DiffRas}}$ (b) without $\mathcal{L}_{\mathrm{face}}$ (c) without $\mathcal{L}_{\mathrm{eik}}$ and (d) without $\mathcal{L}_{\mathrm{delta}}$}
\label{fig:ablation}
\end{figure}

We evaluate our results with datasets commonly used for learning full body dynamic avatars from videos and compare results with state-of-the-art baselines. 
\subsection{Dataset}

\noindent\textbf{ZJU MoCap}~\cite{peng2021neural}: The ZJU MoCap dataset consists of actors performing various activities captured in 23 videos. In all our settings, we train our model on 4 views (as in~\cite{peng2021neural}) on 300 frames. We train the multi-identity model on 4 simultaneous identities. We evaluate novel view information on the remaining cameras of training poses and novel pose information on test cameras of unseen poses. 

\noindent\textbf{Human3.6M}~\cite{h36m_pami}: We follow the protocol defined in\cite{peng2021animatable} and train our model on 3 cameras and evaluate the performance of our method on the remaining cameras for novel view and poses on a selected subset of of Human3.6M dataset.

\subsection{Baselines}
We compare our approach with the following baselines:\\
\noindent\textbf{Animatable NeRF}~\cite{peng2021animatable}. Canonicalizes sampled points based on the skinning weights of a parametric mesh and learns a single body model for each data. The main difference of \cite{peng2021animatable} and our approach involves the residual calculation. Particularly, animatable-nerf learns residual blend weights rather than a residual in canonical space as in our framework. \\
\noindent\textbf{NeuralBody}~\cite{peng2021neural}. NeuraBody learns to associate a latent code with each vertex of a deformable body model and then performs ray tracing on the diffused version of these learned latent codes. \\
\noindent\textbf{A-NeRF}~\cite{su2021anerf}. Uses only pose information and defines pose relative encodings to learn an avatar that can be rendered in novel views and poses. \\
\noindent\textbf{NHR}~\cite{wu2020multi}. Neural Human Rendering uses sparse point clouds to render multi-view images without explicit pose based reasoning.
\subsection{Quantitative Analysis}
We evaluate the approaches using MSE, PSNR, SSIM and LPIPS metrics for all testing frames in all settings. Table~\ref{tab:comparison} and Table~\ref{tab:comparisonh36} compare our approach with the state-of-the-art methods on ZJU and Human3.6M respectively. We use S9 and S11 for evaluation for Human3.6M, which are the commonly used testing subjects. The generated images used for evaluation of the state-of-the-art methods are provided by~\cite{peng2021animatable}.

\subsection{Qualitative Analysis}

Fig.~\ref{fig:zjuc} and Fig.~\ref{fig:comparison} show the performance of our method compared to above methods on the ZJU-MoCap and Human3.6 dataset, respectively. We observe that Animatable-NeRF performs poorly on novel poses without additional refinement. Inference time refinement is inconvenient as it involves an optimization step which adds some cost to the inference speeds. Additionally, as seen in the last row of Fig.~\ref{fig:comparison}, Animatable-NeRF fails to maintain the shape of the actor under novel poses. NeuralBody generates high quality results for novel views, however struggles with novel poses. 

Additionally, We found that A-NeRF overfits to training views due to its high dimensional pose relative encoding which require large number of poses to train. In contrast, our c-NeRF renderer is able to generalize to novel poses, due to pixel level conditioning information from intermediate neural features. Fig~\ref{fig:anerf} shows the intermediate output of the DiffRas stage and the reconstructed geometry for the novel poses. We see that the c-NeRF adds additional texture information while also providing access to detailed geometry of the actor.
\vspace{-5mm}
\subsection{User study}
Since most of the metrics used serve mostly as a proxy to perceptual quality, we perform a user study to compare avatars generated using different methods. Particularly, each user is presented with 20 randomly sampled views in novel poses across 4 different identities for 8 seconds. The user is asked to rate which of the 4 avatars looks most realistic. We collected this data over 35 random subjects. The demographics include both technical experts who work on digital avatars and general audience. The results (Table.~\ref{tab:comparison}) of the study indicate that our method is preferred $\textbf{66 \%}$ of the time as compared to the the next highest rated method (NeuralBody \cite{peng2021neural}) at $24 \%$.

\begin{table}[h]
\footnotesize
    \centering
    \begin{tabular}{l|ccccc}
     & MSE  $\downarrow$ & SSIM $\uparrow$ & PSNR $\uparrow$ & \multicolumn{2}{c}{LPIPS $\downarrow$} \\
     &&&& Alexnet & VGG \\
     \specialrule{0.12em}{0.05em}{0.05em}
    A-NeRF~\cite{su2021anerf}  & $2.34$ & $0.946$ & $26.47$	& $0.082$ &	$0.058$ \\
    NHR~\cite{wu2020multi} & $0.82$ &	$0.976$ & $27.91$ & $0.026$ & $0.024$ \\
    NeuralBody~\cite{peng2021neural} & $0.66$ & $0.977$ & $27.93$ & $0.039$ &	$0.029$ \\
    Anim-NeRF~\cite{peng2021animatable} & $0.50$ &	$0.979$ &	$28.11$ &	$0.035$ &	$0.028$ \\ 
    \hline
    Ours & $\mathbf{0.32}$	& $\mathbf{0.985}$ & $\mathbf{29.42}$ & $\mathbf{0.023}$ & $\mathbf{0.022}$
    \end{tabular}
    \caption{Novel pose synthesis results on Human3.6M dataset. Our approach outperforms recent baselines on most supervised metrics.}
    \label{tab:comparisonh36}
\end{table}

\subsection{Ablation Study}
We study the following important questions with respect to our framework.

\noindent\textbf{Need for DiffRas Objective.}
The DiffRas objective constrains the first four channels of the intermediate features, we see in Fig.~\ref{fig:ablation} and Table~\ref{tab:ablation} that without this constraint, the c-NeRF module cannot effectively learn higher frequency information and the results are significantly noisier. This constraint helps ease the learning problem by providing supervision for lower frequency information (since it is trained at a lower resolution).\\

\noindent \textbf{Effect of Face Regularization.}
The advantage of the SDF representation is seen by the use of this regularization. Since tracking errors can lead to significant facial blurring, regularizing the face shape w.r.t base SMPL mesh alleviates the texture averaging in facial regions.\\
\begin{table}[h]
\small
    \centering
    \begin{tabular}{c|cccc}
     & MSE  $\downarrow$ & SSIM $\uparrow$ & PSNR $\uparrow$ & LPIPS $\downarrow$ \\
     \specialrule{0.12em}{0.05em}{0.05em}
    Ours($-\mathcal{L}_{\mathrm{DiffRas}}$)  & $3.511$ & $0.957$ & $24.546$ & $0.053$ \\\
    Ours($-\mathcal{L}_{\mathrm{eik}}$)    & $1.767$ & $0.975$ & $27.526$ & $0.033$    \\
    Ours($-\mathcal{L}_{\mathrm{face}}$)   & $1.785$ & $0.975$ & $27.485$ & $0.031$    \\
    Ours($-\mathcal{L}_{\mathrm{delta}}$) & $\mbf{1.402}$ & $0.977$ & $27.531$ & $0.031$   \\
    \hline
    Ours & $1.568$ & $\mbf{0.977}$ & $\mbf{28.045}$ & $\mbf{0.026}$  \\
    \end{tabular}
    \caption{Ablation results on models trained on ID-337 of ZJU-MoCap dataset.}
    \label{tab:ablation}
    \vspace{-4mm}
\end{table}
\vspace*{-3mm}
\section{Discussion}
\noindent\textbf{Limitations and Future Work.}
Although our approach is robust to large pose or view changes, tracking errors during parametric body fitting can create inaccurate canonicalization or misalignments in neural textures. It may be fruitful to explore directions which do not require explicit parametric model fitting~\cite{saito2021scanimate} or learn 3D surfaces directly from videos~\cite{oechsle2021unisurf,wang2021neus,yariv2021volume}. Also our end-to-end inference time is potentially slow for real-time applications and interesting future directions include scaling the approach to both high-resolution and real-time rendering.

\noindent\textbf{Ethical Considerations.}
A technique to synthesize photorealistic images of people from a monocular video could be misused to create manipulated imagery of real people. Such misuse of media synthesis techniques pose a societal threat. Viable countermeasure solutions include watermarking the model data or output~\cite{2019stegastamp,yu2021artificial}.

\noindent\textbf{Conclusion.}
By combining parametric body models with high-dimensional neural textures and dynamic conditional radiance field network, our approach demonstrates important steps toward the generation of controllable and photorealistic 3D avatars. We believe the avatar creation from a casual video input may lead to the democratization of the high-quality personalized digital avatars.

{\small
\bibliographystyle{ieee_fullname}
\bibliography{egbib}
}

\end{document}